# Field-of-View Extension for Brain Diffusion MRI via Deep Generative Models

**Chenyu Gao,[a] Shunxing Bao,[a] Michael E. Kim,[b] Nancy R. Newlin,[b] Praitayini Kanakaraj,[b] Tianyuan Yao,[b] Gaurav Rudravaram,[a] Yuankai Huo,[b] Daniel Moyer,[b] Kurt Schilling,[c] Walter A. Kukull,[d] Arthur W. Toga,[g] Derek B. Archer,[h,i] Timothy J. Hohman,[h,i] Bennett A. Landman[a,b,c], Zhiyuan Li*[a]**

[a]Department of Electrical and Computer Engineering, Vanderbilt University, Nashville, TN, USA
[b]Department of Computer Science, Vanderbilt University, Nashville, TN, USA
[c]Department of Radiology & Radiological Sciences, Vanderbilt University Medical Center, Nashville, TN, USA
[d]Department of Epidemiology, University of Washington, Seattle, Washington, USA
[e]Wisconsin Alzheimer's Disease Research Center, University of Wisconsin School of Medicine and Public Health, Madison, WI, USA
[f]Department of Medicine, University of Wisconsin School of Medicine and Public Health, Madison, WI, USA
[g]Laboratory of Neuro Imaging, Stevens Neuroimaging and Informatics Institute, Keck School of Medicine, University of Southern California, Los Angeles, CA, USA
[h]Vanderbilt Memory & Alzheimer's Center, Vanderbilt University Medical Center, Nashville, TN USA
[i]Vanderbilt Genetics Institute, Vanderbilt University Medical Center, Nashville, TN USA

**Abstract**

**Purpose:** In brain diffusion MRI (dMRI), the volumetric and bundle analyses of whole-brain tissue microstructure and connectivity can be severely impeded by an incomplete field-of-view (FOV). This work aims to develop a method for imputing the missing slices directly from existing dMRI scans with an incomplete FOV. We hypothesize that the imputed image with complete FOV can improve the whole-brain tractography for corrupted data with incomplete FOV. Therefore, our approach provides a desirable alternative to discarding the valuable brain dMRI data, enabling subsequent tractography analyses that would otherwise be challenging or unattainable with corrupted data.

**Approach:** We propose a framework based on a deep generative model that estimates the absent brain regions in dMRI scans with incomplete FOV. The model is capable of learning both the diffusion characteristics in diffusion-weighted images (DWI) and the anatomical features evident in the corresponding structural images for efficiently imputing missing slices of DWI in the incomplete part of the FOV.

**Results:** For evaluating the imputed slices, on the WRAP dataset the proposed framework achieved $\text{PSNR}_{b0}$=22.397, $\text{SSIM}_{b0}$=0.905, $\text{PSNR}_{b1300}$=22.479, $\text{SSIM}_{b1300}$=0.893; on the NACC dataset it achieved $\text{PSNR}_{b0}$=21.304, $\text{SSIM}_{b0}$=0.892, $\text{PSNR}_{b1300}$=21.599, $\text{SSIM}_{b1300}$= 0.877. The proposed framework improved the tractography accuracy, as demonstrated by an increased average Dice score for 72 tracts ($p < 0.001$) on both the WRAP and NACC datasets.

**Conclusions:** Results suggest that the proposed framework achieved sufficient imputation performance in brain dMRI data with incomplete FOV for improving whole-brain tractography, thereby repairing the corrupted data. Our approach achieved more accurate whole-brain tractography results with extended and complete FOV and reduced the uncertainty when analyzing bundles associated with Alzheimer's Disease.

***Corresponding author, E-mail:** zhiyuan.li@vanderbilt.edu

# 1 Introduction

Diffusion magnetic resonance imaging (dMRI) offers a non-invasive, in vivo approach for measuring the diffusion of water molecules in biological tissues and has become a well-established technique for studying human white matter microstructure and connectivity[1–4]. The movement of water molecules is often restricted by biological structures such as cell membranes and axonal fibers, resulting in a preferred direction of movement that reflects the properties of tissues. A standard dMRI scan is designed to acquire multiple volumes under varying magnetic fields (i.e., by applying diffusion-encoding magnetic gradient pulse from a number of non-collinear directions), such that each volume selectively captures the propensity of water diffusivity in a particular direction, thereby yielding diffusion-weighted images (DWIs). The effect of the gradient pulse, both in terms of time and strength, is characterized by a parameter known as the b-value. Additionally, the orientation of the gradient is commonly specified as a unit-length vector known as the b-vector, and the high diffusivity of water molecules along the gradient orientation yields high signal attenuation. Reference volumes with no diffusion signal attenuation, i.e., with a b-value equals to $0 \text{ s/mm}^2$ and a b-vector equals to $(0,0,0)$, are also required to be acquired during dMRI scan and are often referred as b0 images. To quantify the properties of water diffusion in brain tissues, voxel-wise scalar metrics like mean diffusivity (MD), and fractional anisotropy (FA) are derived from an assumed diffusion tensor (ellipsoidal) model[5]. Additionally, to study whole-brain physical connections, fiber tractography methods delineate the white matter fiber pathways connecting regions of the brain[6,7]. In the last decade, dMRI and its related diffusion measures have become the method of choice to study brain tissue properties and changes associated with Alzheimer's disease, stroke, schizophrenia, and aging[6,8–11].

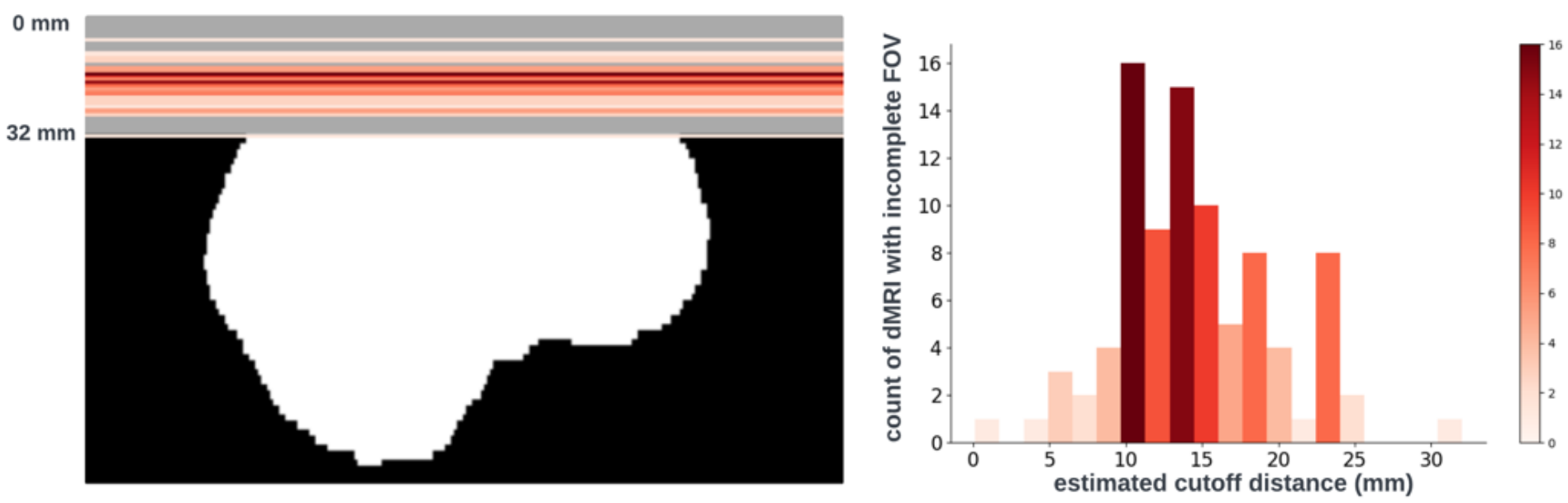

**Figure 1. Visualization (left) and histogram (right) of 103 real cases of dMRI scans with incomplete FOV that failed quality assurance. In the left figure, horizontal red lines and background gray areas indicate where the reduced FOV ends and its corresponding missing regions, respectively, with the estimated position of a brain mask. The total cutoff distance from the reduced FOV to the top of the brain is estimated using a corresponding and registered T1w image.**

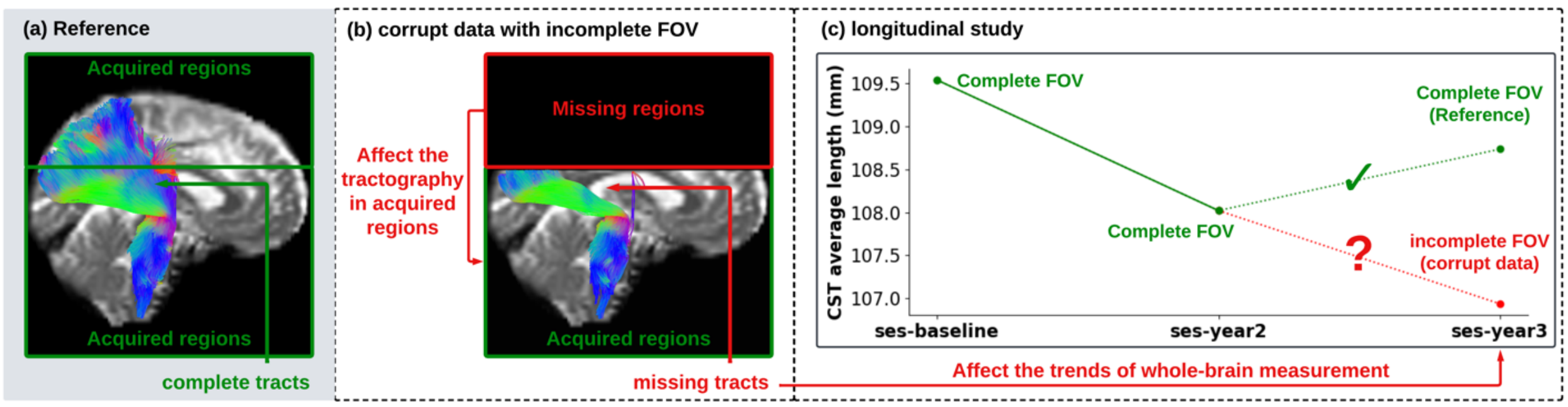

**Figure 2. The missing regions resulting from an incomplete FOV not only render analyses of those areas impossible but can also impact the tractography performed in the acquired regions (as shown in panel (b)), e.g., yielding missing streamlines of corticospinal tract (CST) compared with reference (as shown in panel (a)). Furthermore, whole-brain measurements derived from corrupted data can lead to incorrect interpretations in longitudinal studies (as shown in panel (c)): the measurement from corrupted data (represented by the red dot for the "year3" session) might suggest that the average length of the CST for this subject continues to decrease. This, however, may contradict the fact that when considering correct measurements (represented by green dots).**

Despite the unique clinical capabilities and potential, the whole-brain volumetric and tractography analyses brought by dMRI can be severely impeded by an incomplete field-of-view (FOV), commonly caused by patient misalignment, suboptimal scan plan selection, or necessity in protocol design. A major limitation of dMRI is the extended acquisition time compared with traditional structural MRI, due to the acquisition of volumes with varying diffusion-encoding gradient directions. Typically, protocols with more than 31 directions are recommended for longitudinal studies of disease progress or treatment effects[12]. The long acquisition time further amplifies clinical constraints and imaging artifacts in dMRI like inter-volume motion and eddy-current induced artifacts[13–15]. As a result, the FOV may be incomplete for whole brain scans in suboptimal dMRI acquisition. This then leads to corrupt data with a sequence of brain slices missing in the incomplete part of the FOV, which is one of the most common issues identified during quality assurance of dMRI data[16]. In a recent study of dMRI datasets, we found 103 cases with incomplete FOV out of a total of 1057 cases that failed quality assurance of dMRI preprocessing. The estimated thickness for the missing regions ranged from 1 mm to 32 mm (Figure 1). The loss of information from the missing slices not only prevents analyses in those missing regions but may also affect dMRI-derived analyses of acquired regions (Figure 2), as the global patterns based on the whole brain are impacted. Furthermore, the corrupted data with missing slices introduces bias and inaccuracies for whole-brain analyses, posing significant challenges for longitudinal studies in diagnosing and monitoring neurological developments, including Alzheimer's Disease[17,18].

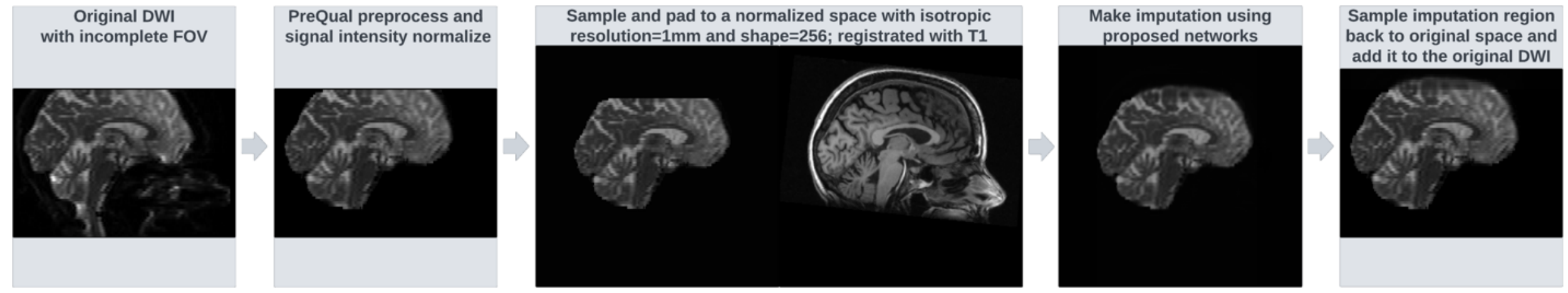

**Figure 3. The pipeline of the proposed FOV extension framework for imputing missing slices in the incomplete part of the FOV begins with PreQual preprocessing and intensity normalization for the DWI in its original space. This is followed by processing the DWI to a normalized space, including resampling and registration with its corresponding T1 image. Subsequently, the proposed 2.5D pix2pix networks are employed to impute the missing slices in the normalized space, utilizing both the DWI (incomplete FOV) and the corresponding T1 (complete FOV). Finally, the imputed regions are resampled back to their original space and added to the original DWI. Sagittal views of a b0 volume with incomplete FOV at each pipeline stage are visualized.**

Since reacquiring the data is not a feasible solution, imputing the missing slices directly from existing scans with an incomplete FOV provides a desirable alternative to discarding the affected but valuable data or re-engineering all downstream methods to accommodate the effects of missing data. Many works have been dedicated to alleviating the impact of missing dMRI data. RESTORE[19] is among the pioneering efforts that introduced an iteratively reweighted least-squares regression for robust estimation of diffusion tensor model by outlier rejection. Recently, TW-BAG[20] developed an inpainting neural network method for repairing the diffusion tensors in cropped regions. For diffusion kurtosis model [21], which further quantifies the non-Gaussianity for water diffusion in brain, REKINDLE[22] proposed a robust estimation procedure to address the increased sensitivity to artifacts and model complexity. However, designing specific methods for each of the numerous and rapidly evolving diffusion and microstructural models would be challenging and inefficient. As an alternative, researchers have also put efforts in repairing the raw DWI signals directly. FSL's *eddy*[23] and SHORE-based method[24] were developed to detect signal dropout and to impute the affected measurements across acquired DWI volumes. However, these methods focus on the imputation of dropout slices based on reference slice signals computed from multiple volumes and cannot be applied to the FOV extension task where no signals are available in the incomplete part of the FOV. A reliable imputation of raw DWI signals for a contiguous sequence of regions in the incomplete part of the FOV remains an unresolved task.

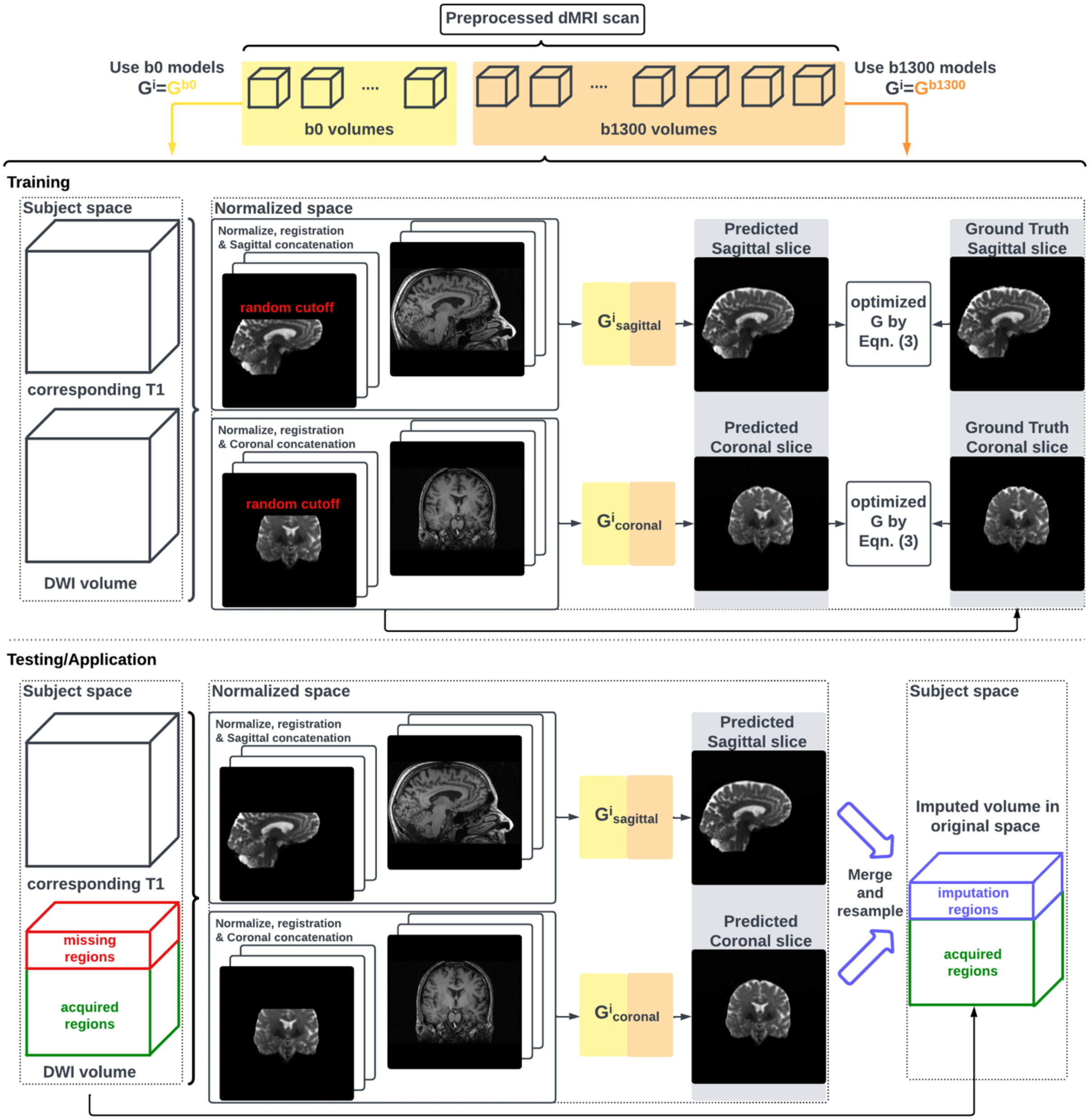

**Figure 4. The whole DWI imputation task is divided into four sub-tasks: imputing b0 volumes' sagittal slices, imputing b0 volumes' coronal slices, imputing b1300 volumes' sagittal slices, and imputing b1300 volumes' coronal slices. The proposed 2.5D networks contain four sub-networks that share the same pix2pix network architecture, and each subnetwork is designed to process a specific sub-task for b0 or b1300 images with their sagittal or coronal slices. During training, random regions are cut-off from either top or bottom of brain to obtain a training DWI data with incomplete FOV, and the subnetworks are optimized to make the imputation of the cut-off regions by Eqn. (3). In testing or application case, for each DWI volume with incomplete FOV, its corresponding sagittal and coronal subnetworks each output an imputed volume by combining every imputed sagittal or coronal slices respectively. These two imputed volumes are then merged into one volume for improving 3D consistency. The imputation regions of the final merged volume are sampled back to the original subject space and are added to the original DWI volume.**

To propose a first solution for this task, we turn to the recent rapid advancements in deep learning, which have shown great potential in image synthesis tasks for dMRI, such as distortion correction[25], denoising[26–28], and registration[29,30]. Directly generating a sequence of dMRI slices in the incomplete part of the FOV, similar to the in-painting task in computer vision, can be challenging, and how to maintain and improve the consistency between the synthesized regions and the observed regions remains an open question[31,32]. Moreover, in medical image synthesis, it is of greater significance for the synthesized regions to conform to the subject's authentic anatomical structures rather than being merely visually realistic. This restrictive requirement of anatomical alignment makes it difficult to naively adapt in-painting models, where the outputs are often diverse[33,34]. However, advantageously, high-quality T1-weighted images are commonly acquired as default alongside a dMRI scan and can be utilized as an anatomical reference. Existing works have shown promising results for integrating the additional anatomical information from T1-weighted images into image synthesis methods for dMRI, such as correcting diffusion distortion by synthesized b0 image[25], synthesizing high angular resolution dMRI data[35], and tractography estimation[36]. Inspired by these findings, in this work we propose a deep generative model framework that imputes the missing brain regions of DWI in the incomplete part of the FOV with extra information from the corresponding T1-weighted image. The proposed model integrates both the diffusion information within the DWI and the structural information of T1-weighted images for accurate imputation of missing slices. A combination of 2.5-dimensional neural networks is proposed for efficient GPU usage and reduced application time. Cross-plane prediction corrections are further applied to improve spatial consistency.

We first train and evaluate our methods on one dMRI dataset with 343 subjects from the same site. To assess generalizability and robustness, we subsequently perform an evaluation on another dMRI dataset with 50 subjects from another site. We reported the missing DWI slice imputation performance using peak signal-to-noise ratio (PSNR) and structural similarity index measure (SSIM). We demonstrate that our approach can improve tractography accuracy for both imputed and acquired brain regions, and to reduce the uncertainty when analyzing bundles associated with Alzheimer's Disease.

The primary contributions of this paper are as follows: (1) We propose a framework that imputes DWI conditioned on T1w images using a deep generative model. We investigate the possibility of synthesizing multi-volume DWI in the incomplete parts of the FOV, as an advancement of existing work that only synthesizes b0 images based on T1w images. The deep generative model fills the gap that traditional imputation methods fail to address, specifically in imputing DWI slices in the incomplete parts of the FOV. (2) We demonstrate that the imputation achieved by our work significantly increases the accuracy of whole brain tractography, thereby repairing corrupted DWI data and make it available for conducting the downstream tract analysis tasks.

## 2 Methods

### *2.1 Problem setting*

Given a diffusion-weighted image $x \in \mathbb{R}^4$ that may have incomplete FOV, we want to learn a mapping from observed image $x$ to output image $y \in \mathbb{R}^4$, $G : x \rightarrow y$, such that $y$ will have complete whole-brain FOV with imputed slices if necessary. To tackle the mapping of DWI with $V$ volumes, we propose to map each volume $x_v \in \mathbb{R}^3$ $(v = 1,2,3, \dots, V)$ separately to its corresponding output volume $y_v \in \mathbb{R}^3$ $(v = 1,2,3, \dots, V)$. Then the output image $y$ is obtained by combining each output volume $y_v$ with the corresponding b-value and b-vector in the gradient table. Directly predicting $y_v$ from $x_v$ can be difficult, given that there are infinite possible gradient directions, each requiring unique feature learning and altogether making the representation learning from $x_v$ complex. We propose to utilize an available T1w image with complete FOV $x_{T1} \in \mathbb{R}^3$ as an extra input, aiming to provide additional information of anatomical structures within $x_{T1}$. Furthermore, given the input pair $\{x_{T1}, x_v\}$, the same $x_{T1}$ shared across all DWI volumes could benefit the optimization of $G$, since it allows the model to leverage a consistent structural reference $x_{T1}$, while learning to predict various missing slices in DWI, focusing on their unique and inherent contrast and directional characteristics within $x_v$. Following the ideas described above, Figure 3 illustrates the comprehensive processing pipeline for the proposed framework of imputing DWI volumes.

| | WRAP | | | NACC | |
|---|---|---|---|---|---|
| | *Alzheimer's Dementia* | *Mild Cognitive Impairment* | *No Cognitive Impairment* | *Dementia* | *Normal* |
| **Train** | 2 | 2 | 241 | N/A | N/A |
| **Val** | 0 | 1 | 48 | N/A | N/A |
| **Test** | 0 | 1 | 48 | 13 | 36 |

**Table 1 Subjects' diagnosis on training, validation, and testing sessions for WRAP and NACC datasets. The names of diagnosis follow the original subject's demographic files.**

## *2.2 Datasets and data preprocessing*

In this study, we initially selected Wisconsin Registry for Alzheimer's Prevention (WRAP)[37] dataset as the primary source for training and evaluating our methodologies. The rationale behind this choice is twofold. Firstly, WRAP dataset contains one of the most extensively corrupted dMRI data in terms of the significant missing regions of brain close to 30mm, due to an incomplete FOV. Secondly, WRAP was collected from a single site, making it an ideal starting point for training and evaluating models without the concerns of variations across multiple sites. Our first cohort on WRAP comprised 343 subjects, each possessing T1w image and single-shell dMRI scans with b-value of $1300\ \mathrm{s/mm^2}$, the most frequent b-value acquired in WRAP. These subjects were split into three distinct groups: 245 subjects for the training set, 49 for the validation set, and 49 for the testing set. Next, to evaluate the robustness and generalizability of the proposed method, we extended our analysis to include the National Alzheimer's Coordinating Center (NACC)[38] dataset that has a large number of dMRI scans sharing the same b-value of $1300\ \mathrm{s/mm^2}$. Our second cohort comprised 49 testing subjects from the same site within NACC, each possessing T1w image and single-shell dMRI scans with b-value of $1300\ \mathrm{s/mm^2}$. Table 1 presents the diagnosis information about the cohorts included in our study.

All DWI were first preprocessed using the PreQual[39] pipeline for correction of susceptibility-induced and eddy-current induced artifacts, slice-wise imputation of mid-brain slices, inter-volume motion, and denoising. Quality assurance checks were performed on PreQual preprocessing reports and output images to ensure valid inputs and successful preprocessing of the data. Next, intensity normalization was performed for each DWI separately, where the maximum value was set to the 99.9th percentile intensity, and the minimum value was set to 0. All volumes of one DWI shared the same normalization parameters. The corresponding T1w image was normalized with a maximum value of its 99.9th percentile intensity and a minimum value of 0. Then the T1w image was registered to DWI by applying an affine transformation computed between the T1w image and the average b0 image of the DWI using FSL's epi reg[40]. Then, both T1w image and all DWI volumes were resampled to $1\text{mm} \times 1\text{mm} \times 1\text{mm}$ resolutions and padded or cropped to $256 \times 256 \times 256$ voxels.

### *2.3 Model*

The proposed neural networks for DWI imputation are presented in Figure 4. For tackling the large GPU memory required by learning the 3D mapping $G : \{x_{T1},\ x_v\} \rightarrow y_v$. We proposed a 2.5D framework to decompose $G$ into two separate generators, $G_{sagittal}$ and $G_{coronal}$ and learn them independently through small patches of 3D volume in sagittal and coronal view respectively. Each small patches contains a sequence of neighboring slices of the target slice ($n$ for each side), and then be used to predict a single slice in sagittal and coronal view. The prediction from sagittal and coronal view are later merged by voxel averaging to obtain the final output volume. We trained separate models to handle the distribution difference between DWI volumes obtained with a b-value equals to $0\ \text{s/mm}^2$ or $1300\ \text{s/mm}^2$, resulting in four generators in total: $G_{b0_sagittal}$, $G_{b0_coronal}$, $G_{b1300_sagittal}$ and $G_{b1300_coronal}$. The axial view was not included in the model because the axial slices of DWI in the incomplete FOV regions is not available, therefore providing no information about the diffusion features for training. We use pix2pix[41] as our generator $G$ for its stable conditional image translation and $L1$ loss for preserving the underlying context of image[42], which is critical for medical image synthesis tasks. The final objective for every $G$ is:

$$\mathcal{L}_{GAN}(G,D) = \mathbb{E}_{y_v}[\log D(y_v)] + \mathbb{E}_{x_v}\left[\log\left(1 - D\left(G(x_{T1}, x_v)\right)\right)\right], \tag{1}$$

$$\mathcal{L}_{L1}(G) = \mathbb{E}_{x_v, y_v}\left[\left|\left|y_v - G(x_{T1}, x_v)\right|\right|_1\right], \tag{2}$$

$$G^* = \text{argmin}_G \max_D \mathcal{L}_{GAN}(G,D) + \lambda \mathcal{L}_{L1}(G), \tag{3}$$

where $D$ is a discriminator to distinguish if the output of generator $G$ looks real.

During training, first a DWI volume and its corresponding T1w image (registered as in data preprocessing) are randomly selected. The DWI volume is randomly cut off by 0 millimeters to 50 millimeters in the normalized space, which covers the maximum missing distance as previously shown in Figure 1 and for model generalizability, from either top or bottom of the brain. The cut-off DWI is then paired with its T1w image as input. The non-cutoff DWI volume is used as ground truth for the prediction. Then, small patches of sagittal and coronal view are created: DWI patches and T1w patches are concatenated along the plane direction. For example, if sagittal DWI patches and T1w patches are both $(2n + 1) \times 256 \times 256$, their concatenation will be $((2n + 1) + (2n + 1)) \times 256 \times 256$. Finally, the corresponding $G_{sagittal}$ and $G_{coronal}$ are optimized by stochastic gradient descent using Eqn. (3), where the expectation of $x_v$ and $y_v$ is approximated by mini-batches of image slices. In our design, we train the model to predict the whole regions of the brain (both cutoff regions and non-cutoff regions) instead of cutoff regions only. We reason that this can encourage the model to learn global representations of the image and thus enhance the model's robustness and generalizability for various size of incomplete FOV, including the case that the input image already has complete FOV. We adopt the state-of-the-art PyTorch implementations (https://github.com/junyanz/pytorch-CycleGAN-and-pix2pix) for training every generator. As suggested in pix2pix, we choose the deterministic $G$ for efficient model training. We used "resnet_9blocks" as the network architecture for $G$ for encouraging the model to explore features within both T1w images and DWI. We set $n = 7$ as the minimum requirement for maintaining 3D consistency. The best model was selected by the imputation performance on the imputed regions only using the validation set.

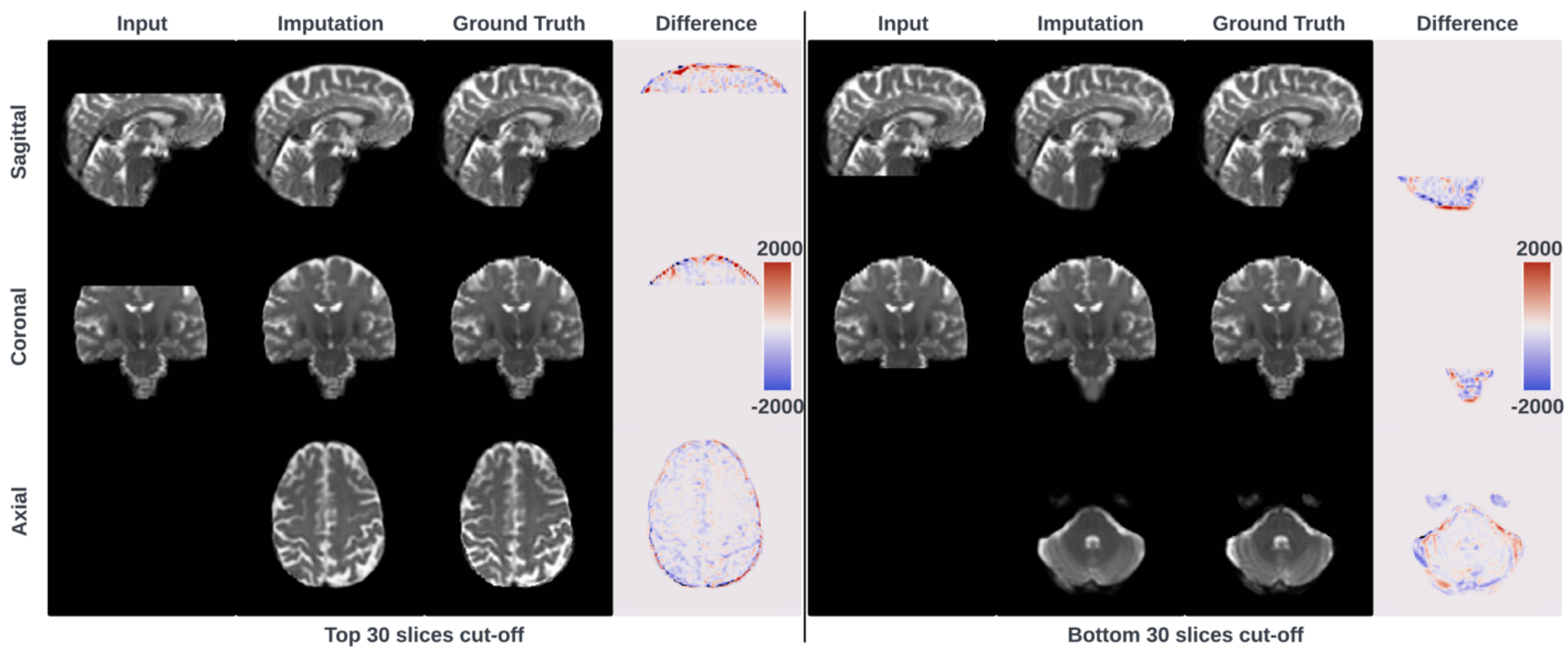

**Figure 5. Imputation for both the top (left panel) and bottom (right panel) of brain. Red and blue indicates that the imputed intensity is larger or smaller respectively than the ground truth. The imputation achieved similar global contrast and anatomical patterns compared to the ground truth reference. Closer examination of local areas, as indicated by the difference image, reveals large imputation errors at the boundaries between white and gray matter, and at the edges of the brain. Additionally, the proposed framework tends to make blurry imputation, thereby losing the high-frequency information that details the brain structure.**

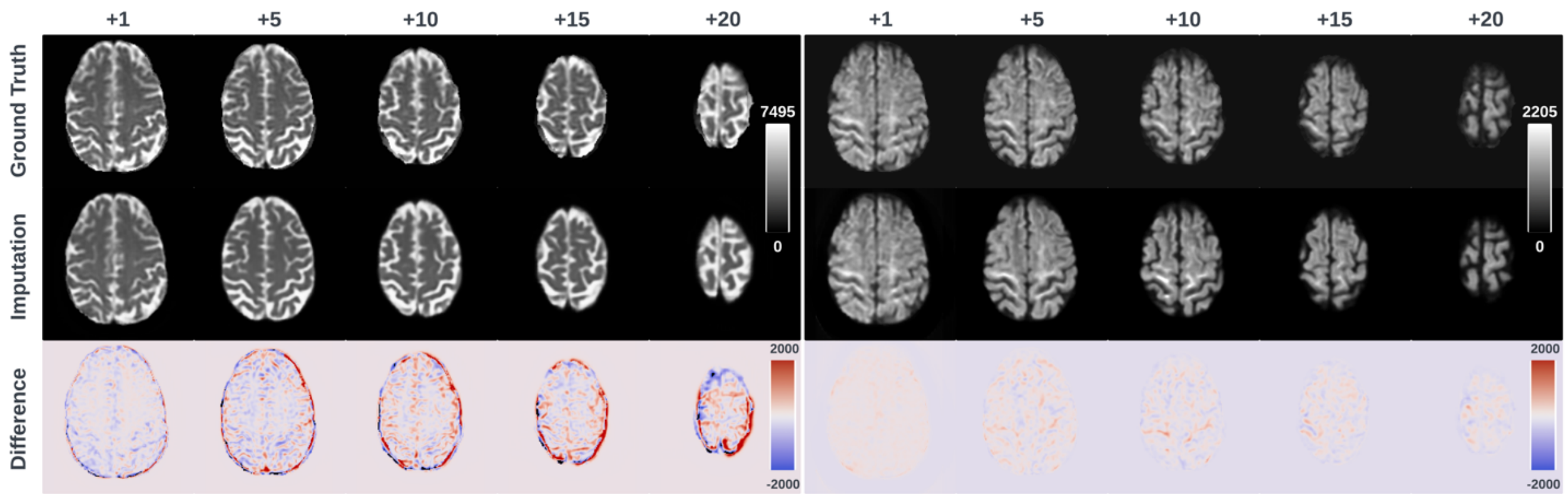

**Figure 6. Axial slice imputations for b0 images (left panel) and b1300 images (right panel). The color lookup tables are adjusted with different intensity ranges for better display of diffusion-weighted volumes. Each column represents the distance to the nearest acquired slice in millimeters (mm). Red and blue indicate that the imputed intensity is larger or smaller respectively than the ground truth reference. Consistent with Figure 5, the proposed framework performs imputations that globally align with the ground truth reference, albeit with a blurrier appearance. Additionally, increasing imputation errors are observed as the distance of the imputed slices increases, for both b0 and b1300 images.**

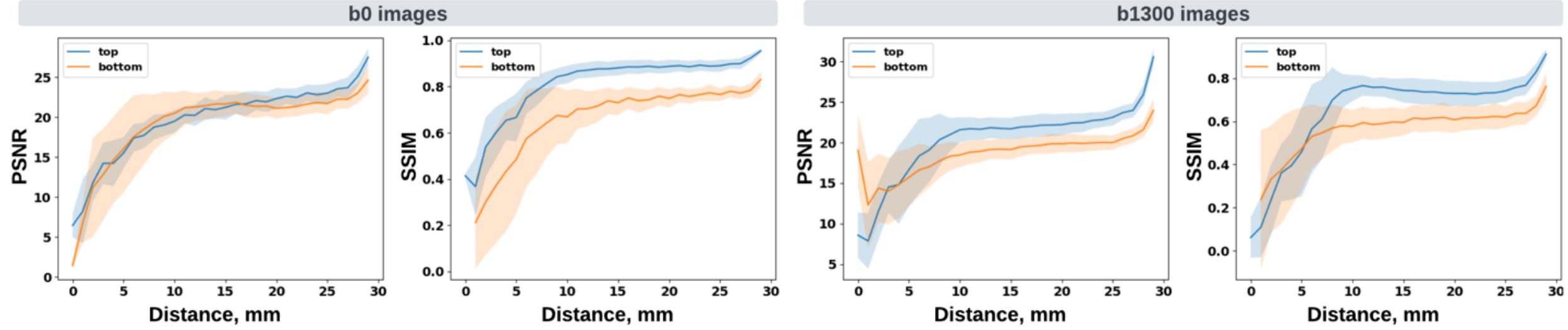

**Figure 7. Imputation performance with respect to the distance from the top or bottom of brain, assuming a complete brain. The larger the distance, the closer to the acquired region. Both PSNR and SSIM metrics for b0 and b1300 images show an ascending trend, indicating an improving imputation accuracy when approaching the nearest acquired region, and a higher error margin in slices adjacent to the top or bottom of brain. At a 30 mm distance, which is approximately the closest missing slice to the acquired brain region, the imputation accuracy markedly improves, as evidenced by the rising tail of each plotted line.**

For testing and application, the model follows the same process to obtain the predicted volume. For the final framework output, we use only the slices in the missing regions of the predicted volume. The imputed regions are sampled back to the original subject space and then combined with the originally acquired regions with incomplete FOV. A mask $m$ that covers the acquired regions (if $m$=1: acquired regions, else: missing regions) can be generated from the testing data with any brain-masking methods ("median_otsu" as a simple example), the final output is therefore $m \odot x_v + (1 - m) \odot \tilde{y}_v$. For all images of the testing subjects, we first cropped them by 30 millimeters to obtain testing images with incomplete FOV. We then used the original full FOV images as our ground truth reference images.

The model is implemented using Python 3.11.5 and PyTorch 2.3.0, along with CUDA 11.8. All experiments are run on an Nvidia Quadro RTX 5000 with 16GB of GPU memory. The batch size is set to 24, and 4 parallel PyTorch data loading workers are used.

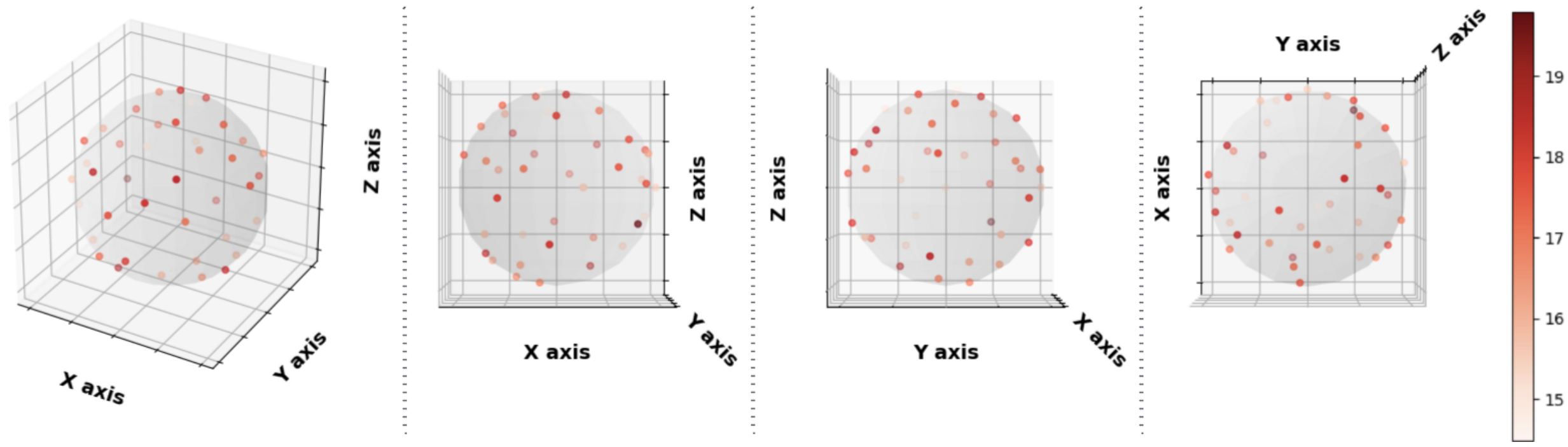

**Figure 8. Imputation performance (PSNR) with respect to 40 directions of diffusion-encoding gradient pulse evaluated by ADC. The average PSNR of ADC is 16.991±1.221. No obvious visual bias is observed for any direction. P-value>0.05 for Kruskal-Wallis test (p=0.999), which fails to reject the null hypothesis that the medians of each direction's measurements are the same.**

### *2.4 Analysis*

First, we qualitatively and quantitatively evaluate the imputation errors on WRAP dataset. We report MSE, PSNR and SSIM for the imputed regions compared with its ground truth reference. The SSIM window are set to 7 for every dimension. Brain masks computed by SLANT-TICV[43] are applied to ensure the metrics are computed for brain areas only. Additionally, we study the imputation performance with respect to the missing slices distance and concerning different directions of diffusion-encoding gradient pulse.

Next, to test our hypothesis that an imputed image with complete FOV, generated by our approach, can improve the whole-brain tractography for corrupted data with incomplete FOV, we conduct paired t-tests for 72 tracts, and specifically investigate 12 of them that are commonly associated with Alzheimer's disease (AD). We present Bland-Altman plots for studying the agreement of bundle shape measurements between the reference and our approach.

Then, to test our hypothesis that T1w images can be helpful for multi-volume dMRI imputation, we conduct an ablation study. This study also serves as a baseline for the proposed model by training a model with the same neural network architecture and settings, but without the input of T1w images.

| | | WRAP | | NACC | |
|---|---|---|---|---|---|
| | | *b0 images* | *b1300 images* | *b0 images* | *b1300 images* |
| **Baseline (no T1w)** | MSE | 45.421±25.106 | 15.843±3.741 | 31.552±17.117 | 8.528±2.071 |
| | PSNR | 16.679±0.976 | 6.458±1.769 | 16.764±1.358 | 7.185±1.606 |
| | SSIM | 0.691±0.066 | 0.255±0.079 | 0.727±0.051 | 0.279±0.071 |
| **Proposed model** | MSE | 12.483±8.041 | 0.418±0.192 | 11.007±5.566 | 0.320±0.141 |
| | PSNR | 22.397±1.573 | 22.479±1.560 | 21.304±1.456 | 21.599±1.299 |
| | SSIM | 0.905±0.047 | 0.893±0.042 | 0.892±0.040 | 0.877±0.021 |

**Table 2. Average MSE, PSNR, and SSIM (3D) for imputation regions of testing data on WRAP and NACC datasets. Our method achieved slightly superior performance on b0 images than on b1300 images, as indicated by the SSIM metrics. The ablation of the T1w image inputs significantly decreases imputation performance, as demonstrated by all three metrics.**

Finally, to evaluate the generalizability of our methods, we report the imputation errors using PSNR and SSIM on an additional NACC dataset. We also conduct the same tractography and bundle analysis on NACC dataset.

## 3 Results

### *3.1 Imputation of Missing Slices*

In general, the proposed method is capable of imputing visually similar slices for both the top and bottom of brain, with similar global contrast and anatomical patterns compared to the ground truth reference. The major differences observed were at the boundaries between white matter and gray matter (Figure 5). The imputation errors increase when the imputed slice is located towards the edges of the brain, i.e., distant from its nearest acquired regions (Figure 6 and Figure 7). MSE, PSNR, and SSIM for the imputed slices of testing subjects are recorded in Table 2. In addition, we studied how the imputation performance can vary in relation to the directions of the diffusion-encoding gradient pulse. The apparent diffusion coefficient (ADC) was computed for 40 directions within the testing subjects. The proposed method showed no obvious bias towards specific directions, as evidenced by the similar PSNR of ADC observed across all directions (Figure 8).

| | WRAP | | NACC | |
|---|---|---|---|---|
| | *Incomplete FOV* | *With imputation* | *Incomplete FOV* | *With imputation* |
| **Acquired Regions** | 0.909±0.026 | 0.933±0.021 | 0.884±0.036 | 0.921±0.022 |
| **Imputed Regions** | N/A | 0.646±0.180 | N/A | 0.643±0.173 |

**Table 3. Average Dice score for 72 tracts produced from image with incomplete FOV and with its imputation. The improvement of imputation over incomplete FOV is statistical significant (p<0.001) from paired t-test on all tracts' results (p=2.52E-20 for WRAP and p=1.25E-24 for NACC).**

### *3.2 Bundle Analyses*

We are interested in how our approach can help repair the bundles and increase the tractography accuracy, within both the acquired regions and the imputed regions. To evaluate this, we ran Tractseg[44] on images with incomplete FOV, their imputed image generated by our approach, and their ground truth reference images with complete FOV. Particularly, we studied a group of 12 tracts, including Rostrum (CC_1), Genu (CC_2), Isthmus (CC_6) and Splenium (CC_7) of the Corpus Callosum (CC) as well as left and right Cingulum (CG), Fornix (FX), Inferior occipito-frontal fascicle (IFO), and Superior longitudinal fascicle I (SLF_I). These tracts are commonly associated with Alzheimer's Disease (AD)[45–59] and were examined for exploring potential clinical benefits of the proposed framework.

As shown in Figure 9, the tracts produced in the imputed regions outside of the previously incomplete FOV are visually very similar to their ground truth reference. However, they lack some streamlines around the edge of the brain. Additionally, in the acquired regions of original DWI, our method improves the accuracy and completeness of tracts that are substantially affected by an incomplete FOV. This improvement is particularly evident in tracts that were previously undetected or only partially produced due the incomplete FOV.

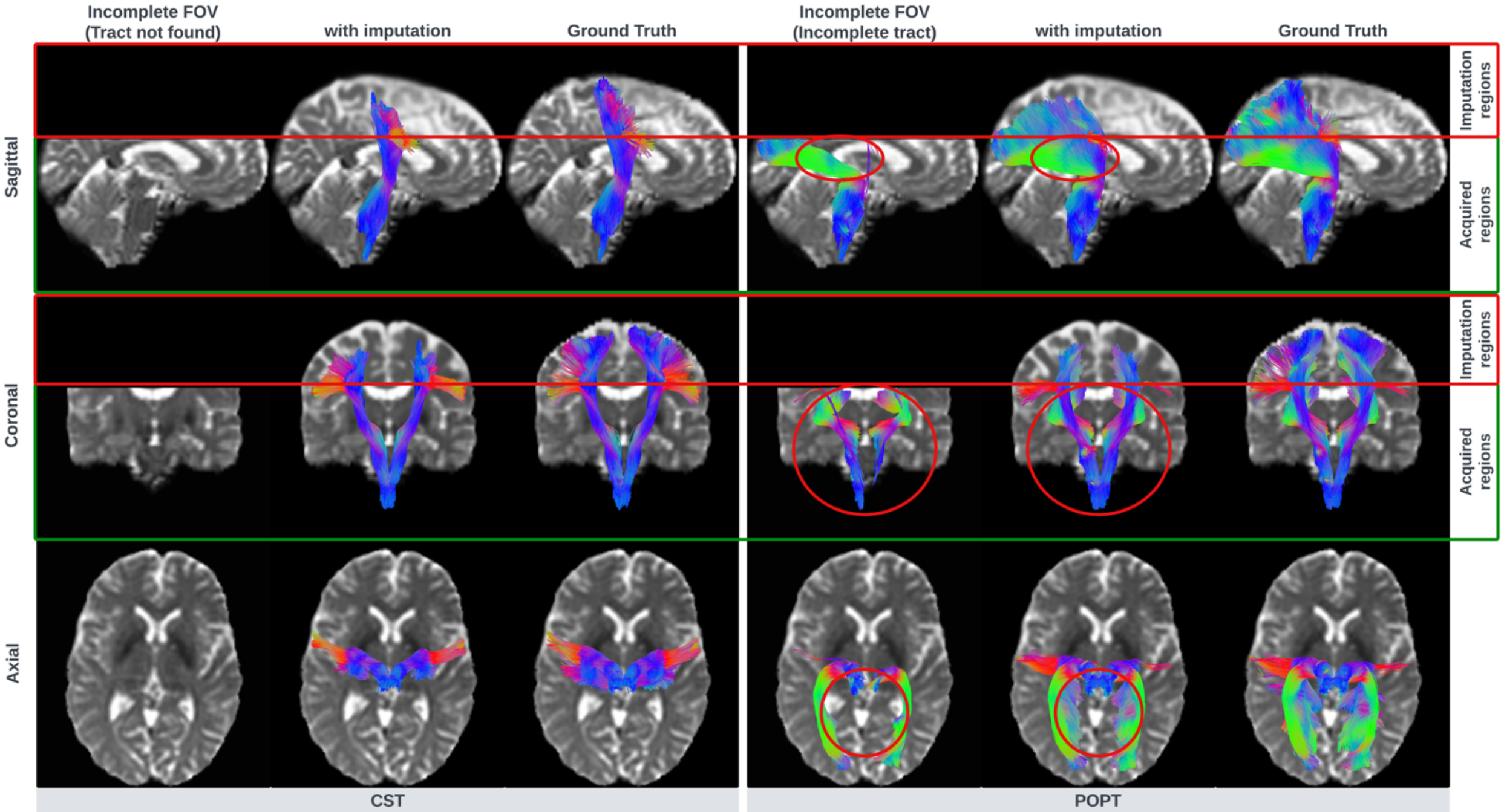

**Figure 9. Tractography results of example tracts for images with incomplete FOV alongside their imputed counterparts and the ground truth references are presented. The tracts produced through imputed images closely resemble the ground truth reference tracts within the acquired regions but lack some streamlines near the brain's edge in the imputed regions. Our approach notably enhances the accuracy and completeness of tracts that are significantly compromised with an incomplete FOV. As shown in the left panel, CST (Corticospinal tract) is completely not detected for image with incomplete FOV, but with imputation CST is produced successfully. In the right panel, the image with incomplete FOV yields a partial POPT (Parieto-occipital pontine) only, yet the imputed image rectifies and completes the tract's overall shape and structure within the acquired regions.**

| | WRAP | | NACC | |
|---|---|---|---|---|
| | *Incomplete FOV* | *With imputation* | *Incomplete FOV* | *With imputation* |
| **Cutoff tracts** | 0.910±0.024 | 0.939±0.010 | 0.887±0.029 | 0.926±0.012 |
| **No-cutoff tracts** | 0.905±0.032 | 0.918±0.0294 | 0.879±0.048 | 0.909±0.034 |

**Table 5. Average Dice score for cutoff tracts whose ground truth tracts are cut-off by incomplete FOV, and no-cutoff tracts whose ground truth tracts are not cut-off by incomplete FOV, in the acquired regions. Our approach can improve tractography accuracy regardless of the ground truth tracts are cut-off by incomplete FOV or not. The improvement is statistical significant ($p<0.001$) for both cutoff tracts ($p=8.41E-17$ for WRAP, $p=6.26E-18$ for NACC), and for no-cutoff tracts ($p=5.76E-12$ for WRAP, $p=2.06E-8$ for NACC).**

Quantitatively, Dice similarity coefficient (Dice score) was computed for all 72 tracts generated by Tractseg. For an accurate comparison, we analyze the tracts derived from images with incomplete FOV alongside those from their corresponding imputed images. Both are matched against the same tract segmentation obtained from the ground truth image with complete FOV. Subsequently, we calculate two Dice scores: one comparing the reference tracts with those from the incomplete FOV images, and another comparing the reference tracts with those from the imputed images. For ease of reference, we label these scores as "Dice for Incomplete FOV" and "Dice for Imputation", respectively. Our approach significantly improved ($p<0.001$) the quality of all 72 tracts on average in the acquired regions, while achieving reasonable Dice scores in imputed regions (Table 3). Likewise, the enhancement of the 12 tracts commonly associated with AD in acquired regions was statistically significant ($p<0.001$), as shown in Table 4. Additionally, we analyzed two distinct groups of tracts. One group contains 50 cutoff tracts, whose ground truth tracts can be cut off by an incomplete FOV, up to 30mm from the top of the brain. The other group includes 22 no-cutoff tracts, whose ground truth tracts are situated far from the top of the brain and, therefore, are not cut off by an incomplete FOV. For both groups, our approach significantly improved the tractography accuracy (Table 5). For a detailed examination, a comprehensive Dice score comparison of all 72 tracts is presented in Figure 10. Our approach brought improvements to nearly every tract, particularly for projection pathways heavily impacted by the absence of the top parts of the brain, such as the Corticospinal tract (CST). Finally, Bland-Altman plots for examining the bundle shape measurements are presented in Figure 11. Our approach demonstrates a much more consistent agreement with the reference compared to measurements obtained from images with incomplete FOV.

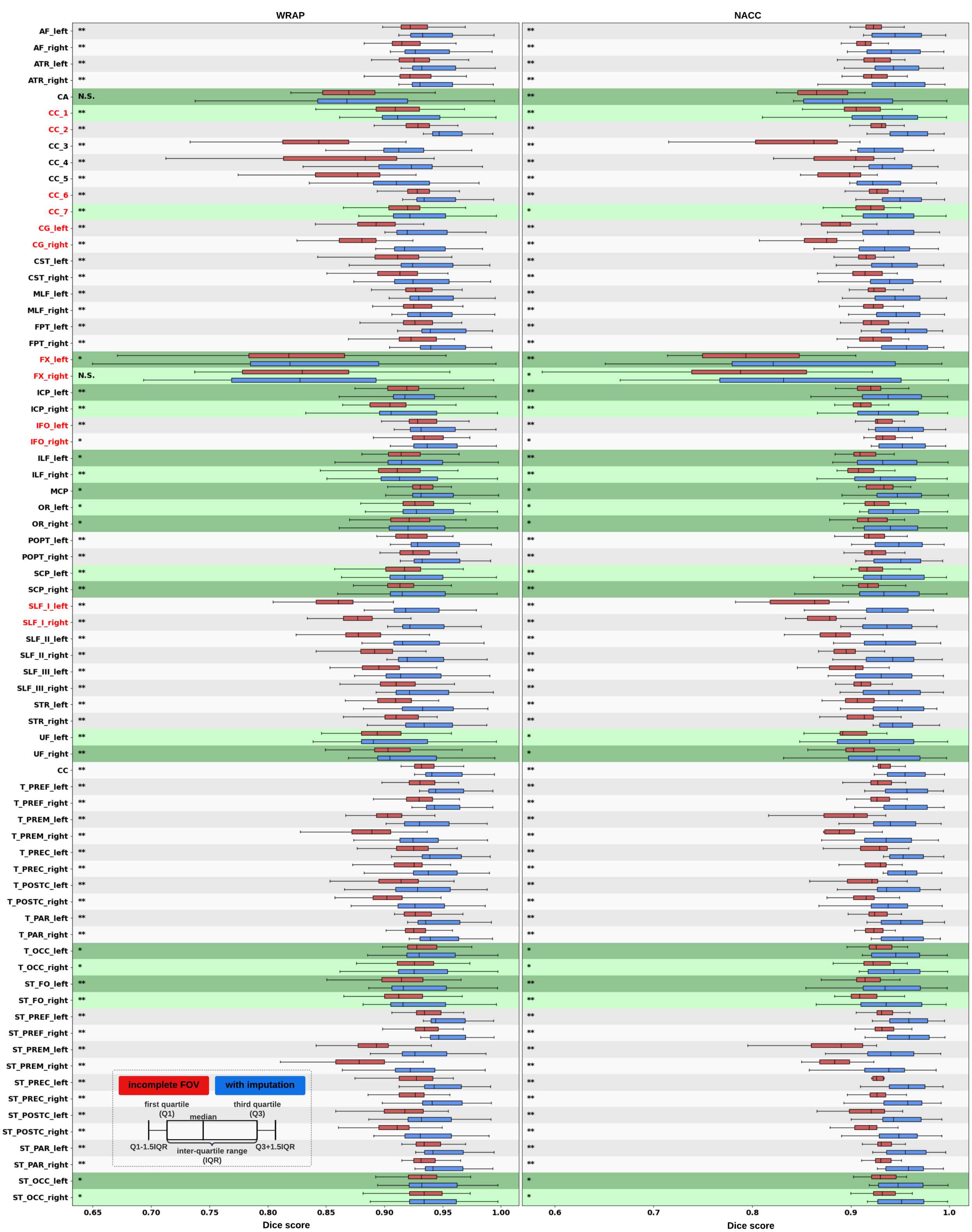

**Figure 10. In both the WRAP and NACC datasets, the proposed framework enhances tractography accuracy through FOV extension (with imputation), as evidenced by the overall higher Dice scores compared to those with an incomplete FOV. Tracts commonly associated with AD have their names in red. Tracts that are not cutoff by incomplete FOV have green shading in their boxplots. Paired t-tests were conducted for each tract and the statistical significance is denoted by "*" ($p < 0.05$), "**" ($p < 0.01$), and "N.S." (not significant).**

# 4 Discussion

In the task of imputing missing DWI slices, our framework exhibited marginally better performance on b0 images compared to b1300 images. This can be attributed to the similarity in patterns between b0 images and T1-weighted images, which makes their joint distribution simpler for the model to learn. This contrasts with b1300 images that require the model to understand additional conditional distributions across various gradient directions. A notable observation was that most imputation errors occurred at the boundary between white matter and gray matter. This is likely because our method tends to predict average intensities over the entire image, which compromises its ability to synthesize sharp intensity contrasts at these areas. Additionally, our method faces greater challenges in imputing slices at the brain's edges. This is evident from the dramatic decreased PSNR and SSIM when the imputed slice is near the top or bottom of the brain. These imputation challenges therefore affect the tractography results, particularly in the difficulties encountered in producing tracts in the same areas.

The comparison of the baseline model with the ablation of T1w image inputs (Table 2), confirms our hypothesis that T1w images are useful for multi-volume DWI imputation. Additionally, we noticed that the performance decreases are much larger for the b1300 images than for the b0 images. This finding supports our motivation that the anatomical information contained in T1w images provides a useful reference for imputing DWI across various directions of water diffusion. It further strengthens the contribution of the proposed framework, which learns to integrate features from both T1w images and multi-volume dMRI scans.

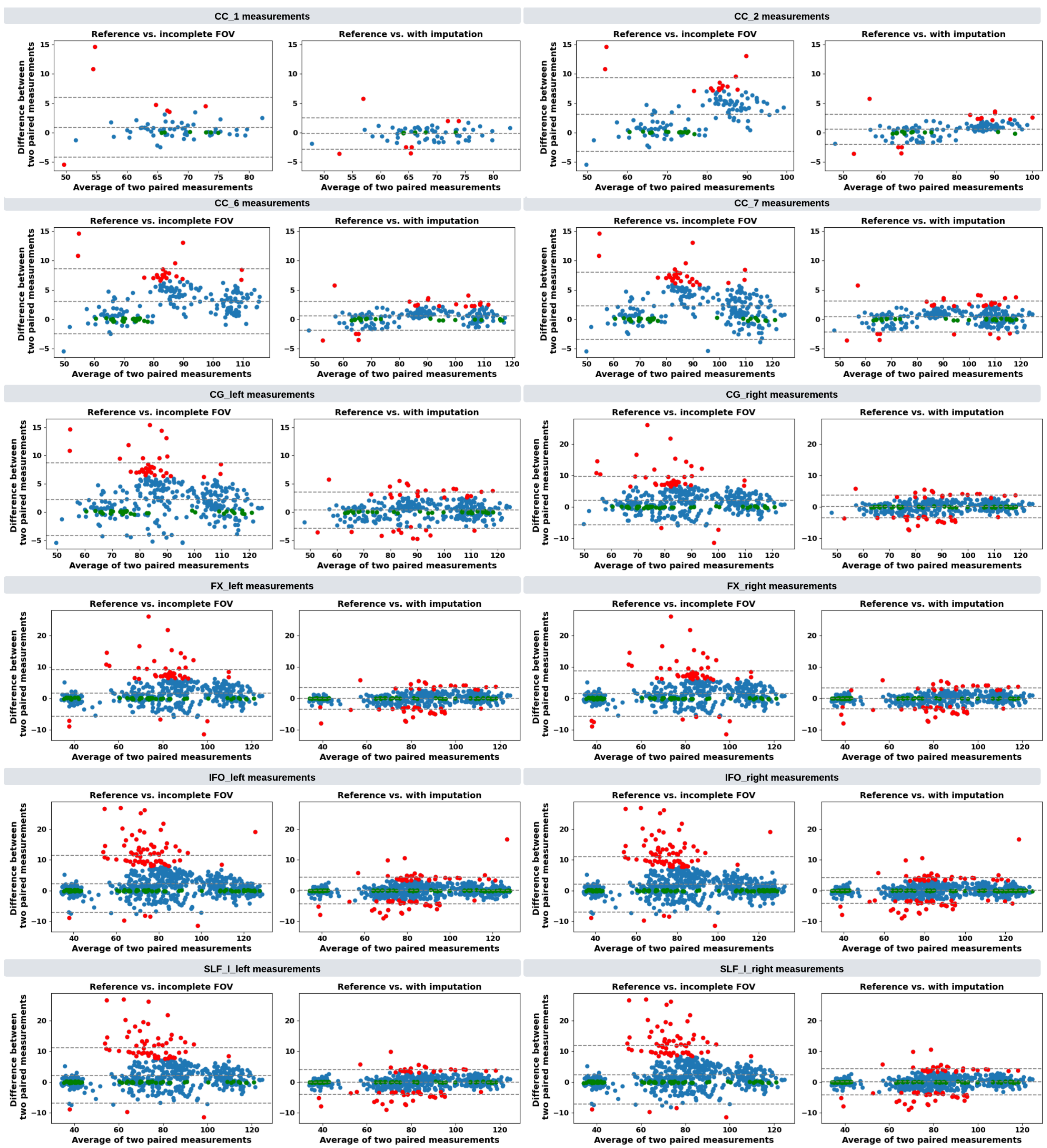

**Figure 11. Bland–Altman plots of agreement for bundle average length compared with reference. The best 10% measurements with the smallest errors are denoted in green, and the worst 10% measurements with the largest errors are denoted in red. Tracts commonly associated with Alzheimer's Disease (AD) that can be impacted by an incomplete FOV (up to 30mm from the top of the brain), specifically CC_1, CC_2, CC_6, CC_7, CG, FX, IFO, and SLF_I are examined. Our approach effectively reduces the significant variations in measurements caused by incomplete FOV. In the "Reference vs. with Imputation" figures, the measurement distribution is tightly clustered and oriented towards the middle dashed-line, indicating coherent and consistent agreement with the reference. In contrast, the "Reference vs. Incomplete FOV" figures show that the measurements span a large range on the y-axis, suggesting substantial errors and variations. By providing consistent measurements of bundles associated with AD, our method can reduce the uncertainty in AD studies that may contain corrupted data due to incomplete FOV.**

It is noteworthy that our approach enhances both the cutoff and no-cutoff tracts. This improvement likely stems from the critical role of whole-brain information in tractography methods. Our findings reinforce the idea that imputing the brain scans in the incomplete part of the FOV can enhance whole-brain tractography and bundle analyses. Consequently, this method holds promise for reducing uncertainty in clinical practice by effectively repairing corrupted data.

## 5 Conclusion

Completing missing dMRI data is a crucial task for the valuable but time-consuming dMRI scans. In this work, we introduced the first method to solve the FOV extension task for DWI. Our framework successfully imputed missing slices in corrupted DWI with incomplete FOV, leveraging information from both diffusion-weighted and T1-weighted images. We evaluated the imputation performance qualitatively and quantitatively on both b0 and b1300 DWI volumes on the WRAP and NACC datasets. The results demonstrated that our model not only effectively imputed the missing DWI slices but also improved subsequent tractography tasks. Most notably, the enhanced accuracy and completeness of tractography and bundle analyses, facilitated by our approach in both imputed and observed regions, underscore the substantial potential in effectively repairing corrupted dMRI data. Future research may focus on advancing the generative model to learn features conditioned on the diffusion signal attenuation ratio $S/S_0$.

## Acknowledgements

This research is supported by NSF CAREER 1452485, NIH 1R01EB017230, NIH NIDDK K01-EB032898, and NIA U24AG074855. This study was supported in part using the resources of the Advanced Computing Center for Research and Education (ACCRE) at Vanderbilt University, Nashville, TN (NIH S10OD023680). We gratefully acknowledge the support of NVIDIA Corporation with the donation of the Quadro RTX 5000 GPU used for this research. The imaging dataset(s) used for this research were obtained with the support of ImageVU, a research resource supported by the Vanderbilt Institute for Clinical and Translational Research (VICTR) and Vanderbilt University Medical Center institutional funding. The VICTR is funded by the National Center for Advancing Translational Sciences (NCATS) Clinical Translational Science Award (CTSA) Program, Award Number 5UL1TR002243-03. The content is solely the responsibility of the authors and does not necessarily represent the official views of the NIH. The ADSP Phenotype Harmonization Consortium (ADSP-PHC) is funded by NIA (U24 AG074855, U01 AG068057 and R01 AG059716). The harmonized cohorts within the ADSP-PHC included in this manuscript were: the National Alzheimer's Coordinating Center (NACC): The NACC database is funded by NIA/NIH Grant U24 AG072122. NACC data are contributed by the NIA-funded ADRCs: P30 AG062429 (PI James Brewer, MD, PhD), P30 AG066468 (PI Oscar Lopez, MD), P30 AG062421 (PI Bradley Hyman, MD, PhD), P30 AG066509 (PI Thomas Grabowski, MD), P30 AG066514 (PI Mary Sano, PhD), P30 AG066530 (PI Helena Chui, MD), P30 AG066507 (PI Marilyn Albert, PhD), P30 AG066444 (PI John Morris, MD), P30 AG066518 (PI Jeffrey Kaye, MD), P30 AG066512 (PI Thomas Wisniewski, MD), P30 AG066462 (PI Scott Small, MD), P30 AG072979 (PI David Wolk, MD), P30 AG072972 (PI Charles DeCarli, MD), P30 AG072976 (PI Andrew Saykin, PsyD), P30 AG072975 (PI David Bennett, MD), P30 AG072978 (PI Neil Kowall, MD), P30 AG072977 (PI Robert Vassar, PhD), P30 AG066519 (PI Frank LaFerla, PhD), P30 AG062677 (PI Ronald Petersen, MD, PhD), P30 AG079280 (PI Eric Reiman, MD), P30 AG062422 (PI Gil Rabinovici, MD), P30 AG066511 (PI Allan Levey, MD, PhD), P30 AG072946 (PI Linda Van

Eldik, PhD), P30 AG062715 (PI Sanjay Asthana, MD, FRCP), P30 AG072973 (PI Russell Swerdlow, MD), P30 AG066506 (PI Todd Golde, MD, PhD), P30 AG066508 (PI Stephen Strittmatter, MD, PhD), P30 AG066515 (PI Victor Henderson, MD, MS), P30 AG072947 (PI Suzanne Craft, PhD), P30 AG072931 (PI Henry Paulson, MD, PhD), P30 AG066546 (PI Sudha Seshadri, MD), P20 AG068024 (PI Erik Roberson, MD, PhD), P20 AG068053 (PI Justin Miller, PhD), P20 AG068077 (PI Gary Rosenberg, MD), P20 AG068082 (PI Angela Jefferson, PhD), P30 AG072958 (PI Heather Whitson, MD), P30 AG072959 (PI James Leverenz, MD); National Institute on Aging Alzheimer's Disease Family Based Study (NIA-AD FBS): U24 AG056270; Religious Orders Study (ROS): P30AG10161,R01AG15819, R01AG42210; Memory and Aging Project (MAP - Rush): R01AG017917, R01AG42210; Minority Aging Research Study (MARS): R01AG22018, R01AG42210; Washington Heights/Inwood Columbia Aging Project (WHICAP): RF1 AG054023; and Wisconsin Registry for Alzheimer's Prevention (WRAP): R01AG027161 and R01AG054047. Additional acknowledgments include the National Institute on Aging Genetics of Alzheimer's Disease Data Storage Site (NIAGADS, U24AG041689) at the University of Pennsylvania, funded by NIA.

**List of Figure Captions**

Figure 1. Visualization (left) and histogram (right) of 103 real cases of dMRI scans with incomplete FOV that failed quality assurance. In the left figure, horizontal red lines and background gray areas indicate where the reduced FOV ends and its corresponding missing regions, respectively, with the estimated position of a brain mask. The total cutoff distance from the reduced FOV to the top of the brain is estimated using a corresponding and registered T1w image.

Figure 2. The missing regions resulting from an incomplete FOV not only render analyses of those areas impossible but can also impact the tractography performed in the acquired regions (as shown in panel (b)), e.g., yielding missing streamlines of corticospinal tract (CST) compared with reference (as shown in panel (a)). Furthermore, whole-brain measurements derived from corrupted data can lead to incorrect interpretations in longitudinal studies (as shown in panel (c)): the measurement from corrupted data (represented by the red dot for the "year3" session) might suggest that the average length of the CST for this subject continues to decrease. This, however, may contradict the fact that when considering correct measurements (represented by green dots).

Figure 3. The pipeline of the proposed FOV extension framework for imputing missing slices in the incomplete part of the FOV begins with PreQual preprocessing and intensity normalization for the DWI in its original space. This is followed by processing the DWI to a normalized space, including resampling and registration with its corresponding T1 image. Subsequently, the proposed 2.5D pix2pix networks are employed to impute the missing slices in the normalized space, utilizing both the DWI (incomplete FOV) and the corresponding T1 (complete FOV). Finally, the imputed regions are resampled back to their original space and added to the original DWI. Sagittal views of a b0 volume with incomplete FOV at each pipeline stage are visualized.

Figure 4. The whole DWI imputation task is divided into four sub-tasks: imputing b0 volumes' sagittal slices, imputing b0 volumes' coronal slices, imputing b1300 volumes' sagittal slices, and imputing b1300 volumes' coronal slices. The proposed 2.5D networks contain four sub-networks that share the same pix2pix network architecture, and each subnetwork is designed to process a specific sub-task for b0 or b1300 images with their sagittal or coronal slices. During training, random regions are cut-off from either top or bottom of brain to obtain a training DWI data with incomplete FOV, and the subnetworks are optimized to make the imputation of the cut-off regions by Eqn. (3). In testing or application case, for each DWI volume with incomplete FOV, its corresponding sagittal and coronal subnetworks each output an imputed volume by combining every imputed sagittal or coronal slices respectively. These two imputed volumes are then merged into one volume for improving 3D consistency. The imputation regions of the final merged volume are sampled back to the original subject space and are added to the original DWI volume.

Figure 5. Imputation for both the top (left panel) and bottom (right panel) of brain. Red and blue indicates that the imputed intensity is larger or smaller respectively than the ground truth. The imputation achieved similar global contrast and anatomical patterns compared to the ground truth reference. Closer examination of local areas, as indicated by the difference image, reveals large imputation errors at the boundaries between white and gray matter, and at the edges of the brain. Additionally, the proposed framework tends to make blurry imputation, thereby losing the high-frequency information that details the brain structure.

Figure 6. Axial slice imputations for b0 images (left panel) and b1300 images (right panel). The color lookup tables are adjusted with different intensity ranges for better display of diffusion-weighted volumes. Each column represents the distance to the nearest acquired slice in millimeters (mm). Red and blue indicate that the imputed intensity is larger or smaller respectively than the ground truth reference. Consistent with Figure 5, the proposed framework performs imputations that globally align with the ground truth reference, albeit with a blurrier appearance. Additionally, increasing imputation errors are observed as the distance of the imputed slices increases, for both b0 and b1300 images.

Figure 7. Imputation performance with respect to the distance from the top or bottom of brain, assuming a complete brain. The larger the distance, the closer to the acquired region. Both PSNR and SSIM metrics for b0 and b1300 images show an ascending trend, indicating an improving imputation accuracy when approaching the nearest acquired region, and a higher error margin in slices adjacent to the top or bottom of brain. At a 30 mm distance, which is approximately the closest missing slice to the acquired brain region, the imputation accuracy markedly improves, as evidenced by the rising tail of each plotted line.

Figure 8. Imputation performance (PSNR) with respect to 40 directions of diffusion-encoding gradient pulse evaluated by ADC. The average PSNR of ADC is 16.991±1.221. No obvious visual bias is observed for any direction. P-value>0.05 for Kruskal-Wallis test (p=0.999), which fails to reject the null hypothesis that the medians of each direction's measurements are the same.

Figure 9. Tractography results of example tracts for images with incomplete FOV alongside their imputed counterparts and the ground truth references are presented. The tracts produced through imputed images closely resemble the ground truth reference tracts within the acquired regions but lack some streamlines near the brain's edge in the imputed regions. Our approach notably enhances the accuracy and completeness of tracts that are significantly compromised with an incomplete FOV. As shown in the left panel, CST (Corticospinal tract) is completely not detected for image with incomplete FOV, but with imputation CST is produced successfully. In the right panel, the image with incomplete FOV yields a partial POPT (Parieto-occipital pontine) only, yet the imputed image rectifies and completes the tract's overall shape and structure within the acquired regions.

Figure 10. In both the WRAP and NACC datasets, the proposed framework enhances tractography accuracy through FOV extension (with imputation), as evidenced by the overall higher Dice scores compared to those with an incomplete FOV. Tracts commonly associated with AD have their names in red. Tracts that are not cutoff by incomplete FOV have green shading in their boxplots. Paired t-tests were conducted for each tract and the statistical significance is denoted by "*" ($p < 0.05$), "**" ($p < 0.01$), and "N.S." (not significant).

Figure 11. Bland–Altman plots of agreement for bundle average length compared with reference. The best 10% measurements with the smallest errors are denoted in green, and the worst 10% measurements with the largest errors are denoted in red. Tracts commonly associated with Alzheimer's Disease (AD) that can be impacted by an incomplete FOV (up to 30mm from the top of the brain), specifically CC_1, CC_2, CC_6, CC_7, CG, FX, IFO, and SLF_I are examined. Our approach effectively reduces the significant variations in measurements caused by incomplete FOV. In the "Reference vs. with Imputation" figures, the measurement distribution is tightly clustered and oriented towards the middle dashed-line, indicating coherent and consistent agreement with the reference. In contrast, the "Reference vs. Incomplete FOV" figures show that the measurements span a large range on the y-axis, suggesting substantial errors and variations. By providing consistent measurements of bundles associated with AD, our method can reduce the uncertainty in AD studies that may contain corrupted data due to incomplete FOV.

## Appendix A

### *A.1 Training Graphs*

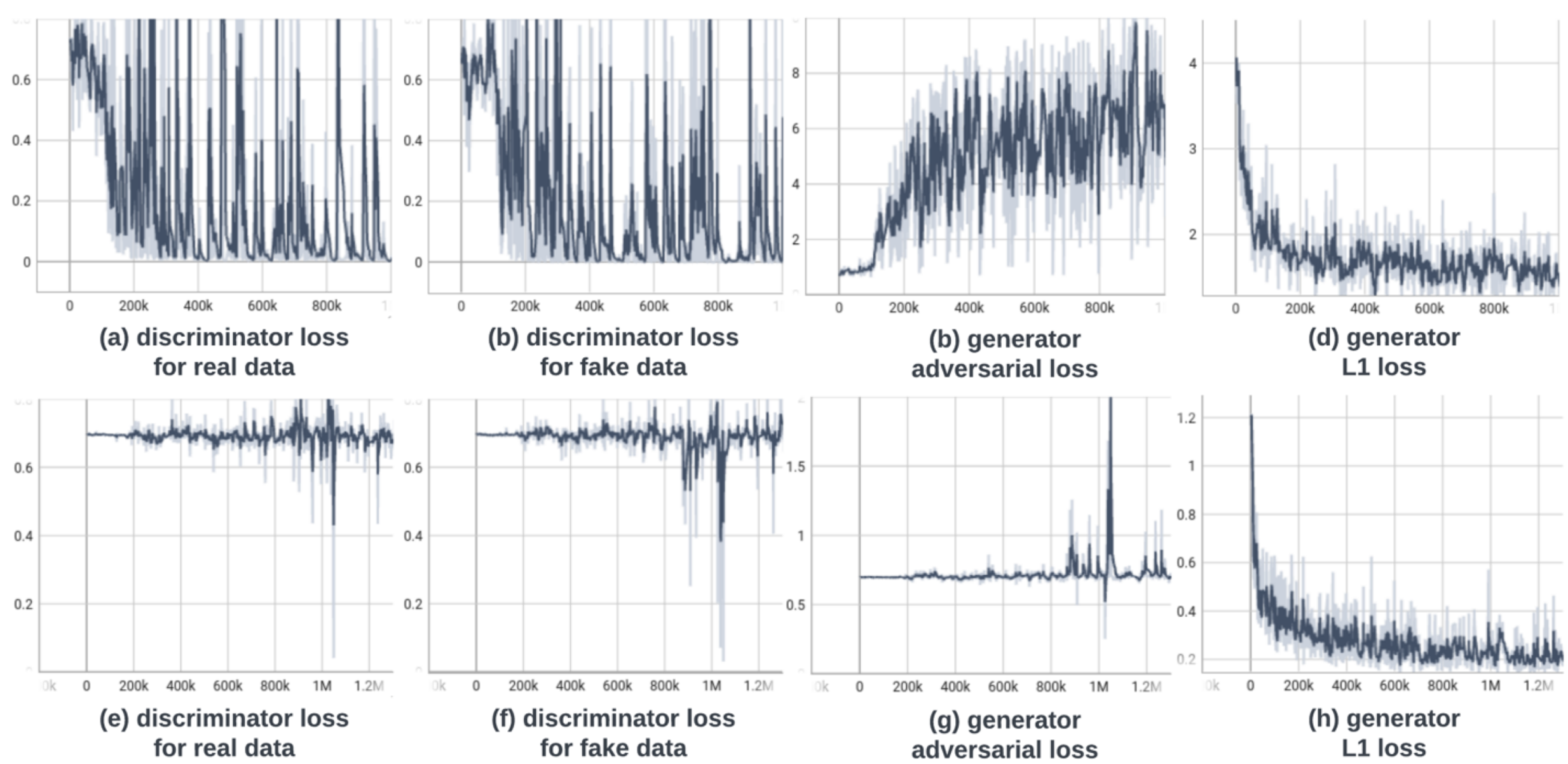

**Figure 12. Training graphs for b0 model (a-d) and b1300 model (e-h). During training, the losses of the discriminator and generator balance each other, demonstrating stable training for their min-max game. The L1 loss of the generator shows a clear decreasing trend and eventually converges.**